\documentclass{article}

\usepackage{arxiv}
\usepackage{xspace}

\usepackage[utf8]{inputenc} 
\usepackage[T1]{fontenc}    
\usepackage{hyperref}       
\usepackage{url}            
\usepackage{booktabs}       
\usepackage{amsfonts}       
\usepackage{nicefrac}       
\usepackage{microtype}      
\usepackage{lipsum}		
\usepackage{graphicx}
\usepackage{natbib}
\usepackage{doi}

\title{Text similarity analysis for evaluation of descriptive answers}

\date{} 				

\author{ \href{https://orcid.org/0000-0003-4998-0073}{\includegraphics[scale=0.06]{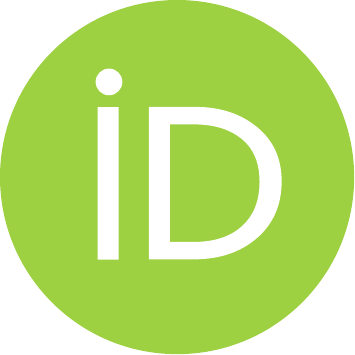}\hspace{1mm}Vedant ~Bahel} \\
	Department of Information Technology\\
	G H Raisoni College of Engineering\\
	Nagpur, India 440016 \\
	\texttt{vbahel@ieee.org} \\
	\And
	\href{https://orcid.org/0000-0002-8425-9962}{\includegraphics[scale=0.06]{orcid.pdf}\hspace{1mm}Achamma~Thomas} \\
	Department of Artificial Intelligence\\
	G H Raisoni College of Engineering\\
	Nagpur, India 440016 \\
	\texttt{achamma.thomas@raisoni.net} \\
}

\renewcommand{\shorttitle}{}

\hypersetup{
pdftitle={Text similarity analysis for evaluation of descriptive answers},
pdfsubject={},
pdfauthor={Vedant ~Bahel, Achamma ~Thomas},
pdfkeywords={Text analysis, Educational Data Mining, Computational Intelligence, MaLSTM},
}

\begin{document}
\maketitle

\begin{abstract}
Keeping in mind the necessity of intelligent system in educational sector, this paper proposes a text analysis based automated approach for automatic evaluation of the descriptive answers in an examination. In particular, the research focuses on the use of intelligent concepts of Natural Language Processing and Data Mining for computer aided examination evaluation system. The paper present an architecture for fair evaluation of answer sheet. In this architecture, the examiner creates a sample answer sheet for given sets of question. By using the concept of text summarization, text semantics and keywords summarization, the final score for each answer is calculated. The text similarity model is based on Siamese Manhattan LSTM (MaLSTM). The results of this research were compared to manually graded assignments and other existing system. This approach was found to be very efficient in order to be implemented in an institution or in an university.
\end{abstract}


\section{Introduction}
In this era, the use of computational intelligence is at a rapidly increasing pace. Every sector is adopting the upcoming technologies in order to advance their traditional systems. The concepts of Machine Learning and Data Science are being implemented in business \citep{larson2016}, finance \citep{kovalerchuk2006data}, manufacturing \citep{bahel2021ci} and healthcare \citep{mane2020computational} as well. There is one such sector which is sometimes neglected, education. Education as a sector holds a large amount of data which includes student’s data, parent’s data, course developer and managements data \citep{romero2007educational}. In \citep{bahel2019}, authors have identified some of the application of Machine Learning and Data Science in Education. In Education sectors, most of the institution are shifting their methods of taking examinations online. But the evaluation of student’s descriptive answers remains manual. Descriptive answers are the long textual answers given against the question. It becomes hefty for a teacher to evaluate dozens of answers of more than dozens of student’s submissions. Sometimes evaluators are often caught being biased towards some students while evaluating the answers.  This paper demonstrates a computational intelligence-based method for automatic evaluation of descriptive answers. Till now the computer aided assessment system can only evaluate the option-based question – answers such as multiple-choice questions or multiple answer questions. The proposed system automatically calculates the score of the answer depending on multiple features including answer correctness, language, grammar and size of the answer. Each feature is weighted differently to provide accurate score to the answer. Such model will be very productive when implemented in universities or institutions. This will ensure faster processing of exam evaluations. Moreover, paper evaluators would be able to take up different efficient task than correcting student’s papers. Authors have used Natural Language Processing (NLP) and certain existing tools to formulate the model \citep{rogers2020can}.
Natural Language Processing is a sub domain of Artificial Intelligence which deals with communication between humans and computer. Chatbots, language translations, smart compose features and speech analysis are all applications of NLP.  Text analysis is one such sub-domain of NLP which deals with analysis of textual data to discover knowledge and insights as per the requirements. The concept of text analysis is derived from various basic concepts like Knowledge Discovery in Databases (KDD), Information retrieval, statistics and general communication .  
The main objective of this research is to use the concepts of text analysis to accurately evaluate descriptive form of answers.

\section{Literature Review}
Researchers have often worked on this since last decade. Earlier researcher majorly focused on developing of better objective answer system. With developing technology, researchers have also started working on auto gradation of descriptive or subjective answers. In \citep{sheeba2014approach}, authors were successfully in deploying a model which could perfectly evaluate single line answers. Most of these existing researches uses keyword analysis as their approach for the model. There are some tools which do not have an extended scope of use and his limited usage opportunity. Mostly the researches focus on short answer grading methodologies \citep{burrows2015eras}. In \citep{kamraj2008survey}, the paper describes Artificial Neural Network based approach to achieve the goal of evaluating descriptive answer. In \citep{biradarsemantic}, authors have used Cosine similarity approach to find the similarity index between two texts. In \citep{kaur2013algorithm}, authors have used Jaccard similarity along with certain word embedding in order to deploy the model. The comparison of performance of models descriped in \citep{kamraj2008survey},\citep{biradarsemantic} and \citep{kaur2013algorithm} is presented in the result and discussion section of this paper. 

\section{Methodology}
\begin{figure}
 \centerline{\includegraphics[width=90mm,scale=1.5]{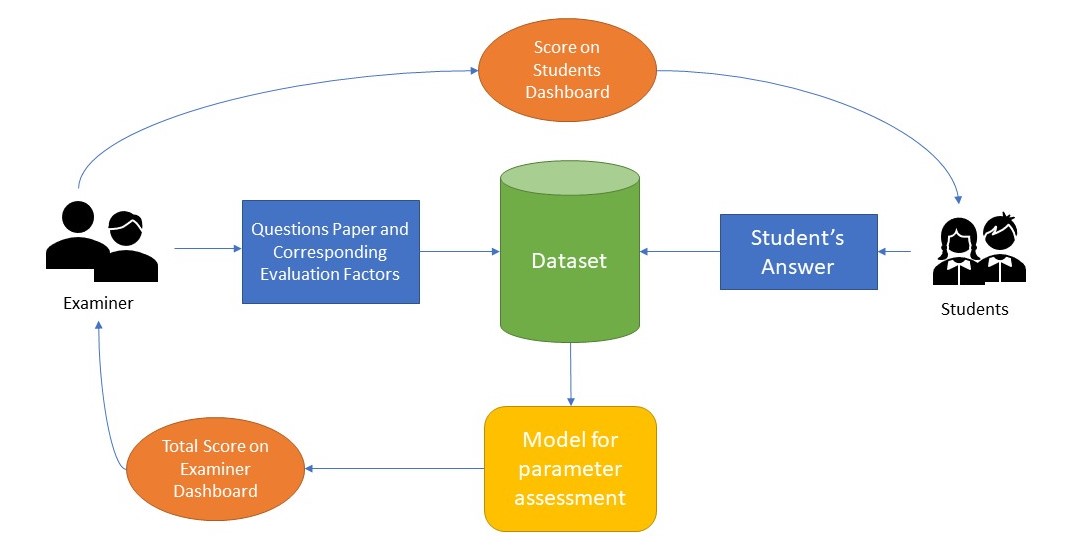}}
	\caption{System Architecture.}
	\label{fig:Fig1}
\end{figure}

The methodology of the proposed system in this paper is explained in reference with a pipeline structure as demonstrated in Fig. \ref{fig:Fig1}.  

\subsection{Examiners Parameters}
As it can be seen in the Fig. \ref{fig:Fig1}, the first step is for the examiner to input the questions and the corresponding evaluation factor. These evaluation factors were chosen as a result of survey conducted at one of the institutions. In this survey, professors were asked for important factors that they consider while evaluating the answer. The input required to evaluate the surveyed factors are as follows:

\subsubsection{Ideal answer}
The examiner needs to input an ideal answer for the given question. Later this answer would be used to for comparison with the students answer in the proposed model.

\subsubsection{Number of works}
Usually most of the questions describe the number of words that the examiner expects as an answer for that question. It is very important that the student answers that question with those word limit in mind.

\subsubsection{Required keywords}
In order to formulate an accurately performing evaluation model, user needs to enter few keywords that are compulsorily expected in the answer by the students.

\subsubsection{Total marks allocated}
Finally, user also needs to assign the total marks allotted for the question. 

\subsection{Scoring parameters}
The above described examiner's parameters along with students’ answers are then loaded into the NLP and text analysis-based intelligent model to calculate the total score. The scoring factor(s) for the model is as follows:

\subsubsection{Size of the answer}
In this model a part of the total score depends upon the size of the answer. The following table demonstrates the marking scheme and corresponding thresholds. The ideal number of words expected for an answer is considered as \textit{x}.
\begin{table}
	\caption{Marking Scheme For Score S1}
	\centering
	\begin{tabular}{lll}
		\toprule
		\multicolumn{2}{c}{}                   \\
		\cmidrule(r){1-2}
		\# of words w.r.t x     & Score (S1) off 1     \\
		\midrule
		Less than  (x-30\%) & 0.5 \\
		Between (x-30\%) \& (x-10\%)    & 0.8\\
		More than  (x+10\%)     & 0.8\\
		x+-10\% & 1 \\
		\bottomrule
	\end{tabular}
	\label{tab:table1}
\end{table}

The score (S1) of this feature contributes to 5\% of the total score for the answer. (Refer Table \ref{tab:table1})

\subsubsection{Language of the answer}
It is very important to check whether the english and grammar used in the answer is proper or not. This model evaluates the structuring and grammar of the answer and gives a score (S2) to it. This score (S2) contributes to 5\% of the total score. Language modelling is an old concept which can be used for this purpose. Language modelling analyses the text and report back with incorrect spellings, incorrect grammar and sentence structures. This is the same concept behind the auto-correct option of various text editors available. The basic concept used behind this is probabilistic methods and text summarization.

\subsubsection{Presence of necessary keywords}
Suppose examiner has listed   number of important keywords. This part of the model checks whether the student has used those keywords in the answer or not. The model not only checks for the existence of those keywords but also checks whether any other word similar in meaning with the keyword is present or not. This simply works on the concept of word bank and simple if-else methods of computation.  If a student has included y1  number of keywords out of y. The score is calculated as follows:

\begin{equation}
	S3= y1/y
\end{equation}

This score (S3) contributes as 10\% in the total score.

\subsubsection{Similarity index of submitted answer and ideal answer}
The most important factor of this whole evaluation system is the similarity index of the submitted answer with the ideal answer as fed by the examiner into the system. It uses various NLP tools and techniques including text semantics analysis. Semantics analysis is a part of NLP which measure the degree of similarity between two pieces of textual data or script. There are multiple algorithms that can be used to achieve this goal. The model described in this paper uses Siamese Manhattan LSTM algorithm (MaLSTM). 

\begin{itemize}
\item First the two pieces of text are converted into individual vectors of numbers through encoding. 
\item These vectors are passed over two LSTM (Long Short-Term Memory) sub networks. LSTM is a Recurrent Neural Network (RNN) approach. These are heavily used for Natural Language Processing. 
\item The semantic meaning of the texts is compared in hidden network and the similarity index is given as output. 
\item The similarity comparison is based on Manhattan distance between two vectors in representation space \citep{mueller2016siamese}. 

\end{itemize}

The score (S4) generated in the form of similarity index from this part of the model contributes to 80\% of the total score.

\subsubsection{Copying index}
Due to shift of examination system online, many universities are facing an issue whether the students are using internet to find the answer. This model does a plagiarism check using the existing open source API’s to govern how much percentage of the content is copies from the external resources. However, this factor doesn’t have a direct impact on the total score but if the copying threshold is more than the permissible threshold, it gives a signal to the examiner that an answer of a particular student is copied. Further, examiner can decide to take whatever action they wish to. 

\begin{table}
	\caption{Scoring Parameters and weight of individual factors on total score}
	\centering
	\begin{tabular}{lll}
		\toprule
		\multicolumn{2}{c}{}                   \\
		\cmidrule(r){1-2}
		Evaluation Factors  & Weight    \\
		\midrule
		Size of the answer (S1) & 5\% \\
		Language of the answer (S2)    & 5\%\\
		Presence of necessary keywords (S3)     & 10\%\\
		Similarity index (S4) & 80\% \\
		Total (St) & 100\% \\
		\bottomrule
	\end{tabular}
	\label{tab:table2}
\end{table}

Then the score St is converted as per the total marks allotted to that answer and the same process is repeated for all the sets of question and answers fed by the examiner (Refer Table \ref{tab:table2}. Once the evaluation process is completed, the score report is sent to the examiner for a quick review. After approval from the examiner, the score is displayed on the student’s dashboard. If in case, the student is not satisfied with the evaluation process, he/she can report back with that specific answer to the examiner who can then evaluated the discrepancy manually in order to have a well-managed examination and evaluation system.

\section{Experiment}
\begin{figure}
 \centerline{\includegraphics[width=90mm,scale=1.5]{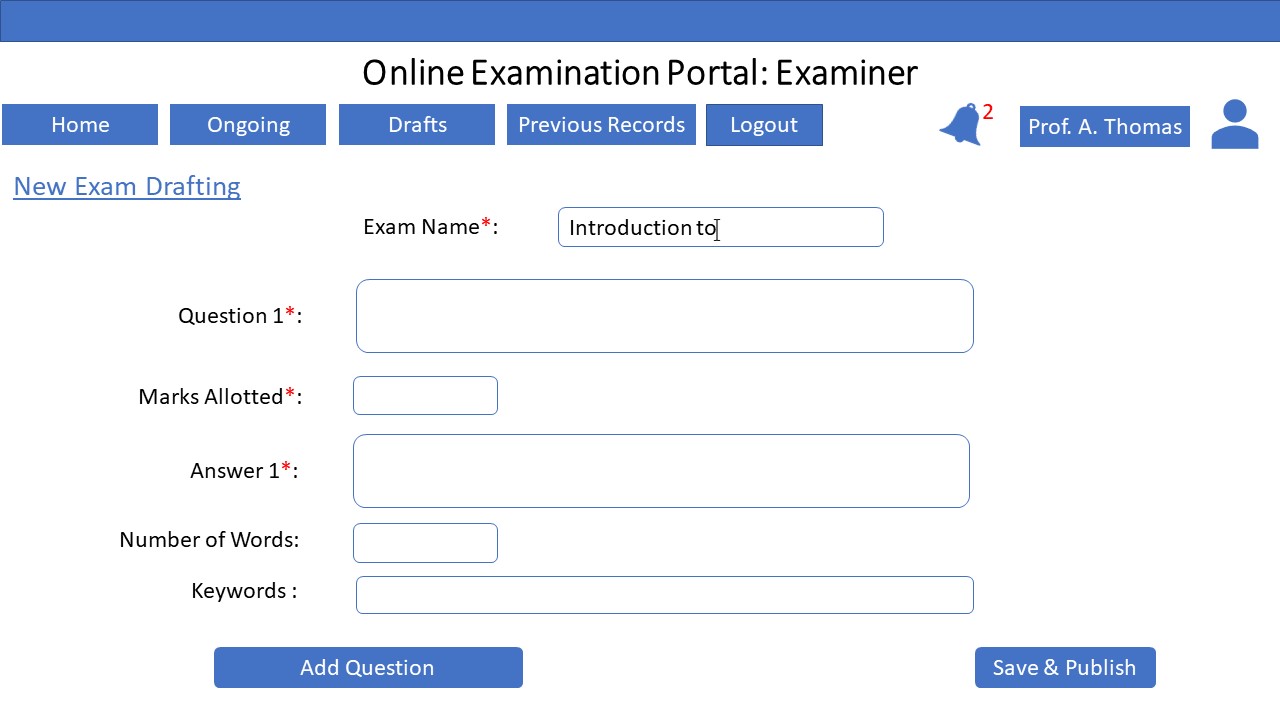}}
	\caption{Examiner’s Dashboard for Online Automatic Exmaination Portal.}
	\label{fig:Fig2}
\end{figure}

In this section, authors have demonstrated the working of the proposed model. Fig. \ref{fig:Fig2} shows the dashboard where the examiner or faculty can input the required information and data for the model to run. Apart from this, the interface also gives option for the examiner to see the previous records, drafted un-published records and user profile. 

\begin{figure}
 \centerline{\includegraphics[width=90mm,scale=1.5]{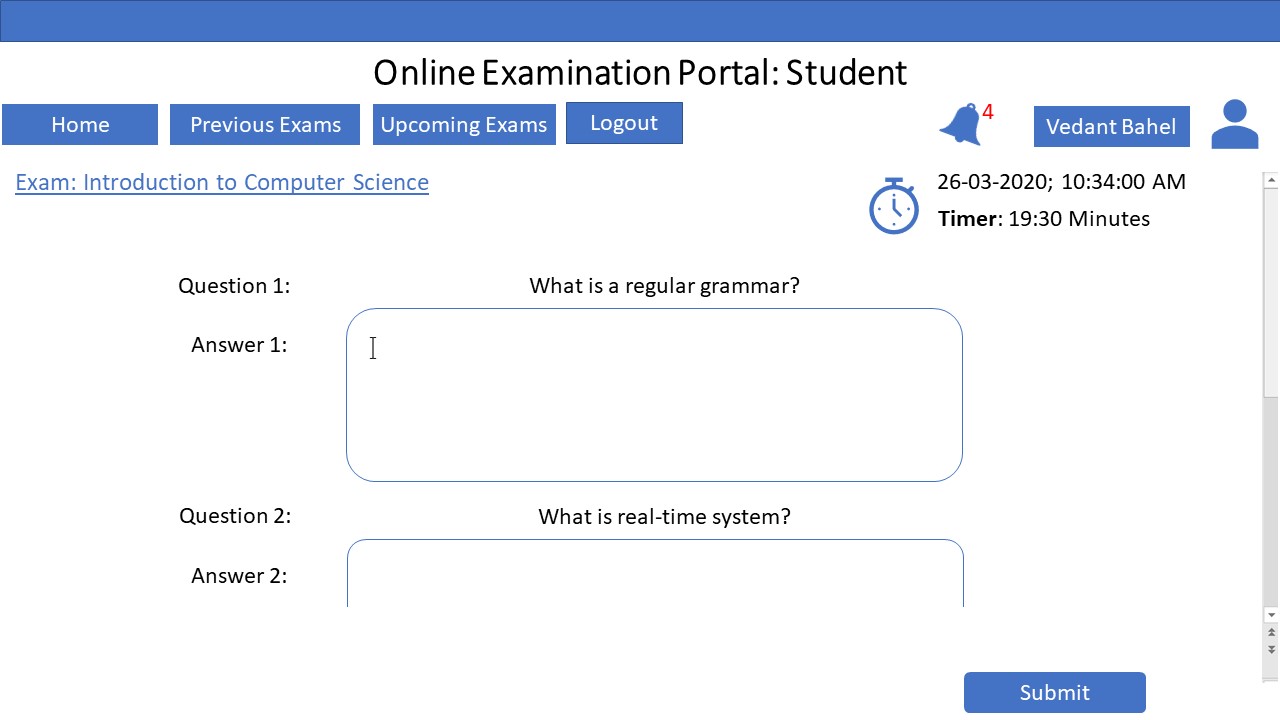}}
	\caption{Student’s Dashboard for Online Automatic Examination Portal.}
	\label{fig:Fig3}
\end{figure}

Similarly, Fig. \ref{fig:Fig3} shows the interface for student to appear for the examination through online window. The dashboard for student also shows previous examination along with the score and an upcoming examination calendar.  

\begin{figure}
 \centerline{\includegraphics[width=90mm,scale=1.5]{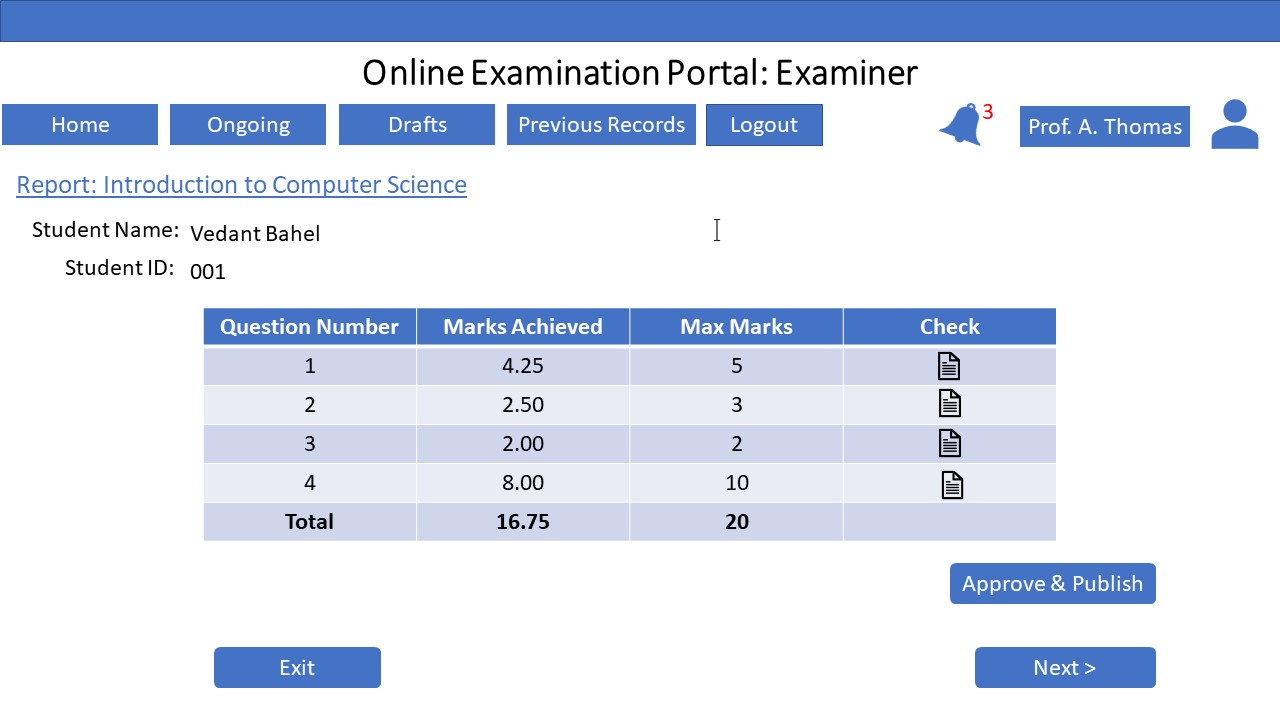}}
	\caption{Exmainer’s Dashboard displaying score of stdudent after evaluation from the model..}
	\label{fig:Fig4}
\end{figure}

Fig. \ref{fig:Fig4} demonstrates the page on the portal where examiner can see the results of the automated model. It shows overall result and student specific results as well. 

The results of the experiment performed on this portal is shown and discussed in the result and discussion part of this paper.  

\section{Result \& Discussion}
To know the performance of this model, authors carried out a simple survey. A question paper was formed, and 4 students were asked to appear for the test. The evaluation for all the four students was done by this model and manually by 3 professors of one of the universities. The result of this model is presented as follows.

\begin{table}
	\caption{Comparison of model performance and manual evaluation}
	\centering
	\begin{tabular}{lllll}
		\toprule
		\multicolumn{5}{c}{}                   \\
		\cmidrule(r){1-5}
		Student     & Auto & Manual & Manual & Manual \\
		ID & Evaluation & Evaluation 1 & Evaluation 2 & Evaluation 3\\
		\midrule
		1 & 17.62 & 17 & 17 & 17 \\
		2 & 16.06 & 14 & 12 & 17 \\
		3 & 18.20 & 18 & 15 & 16 \\
		4 & 17.12 & 16 & 17 & 18 \\
		\bottomrule
	\end{tabular}
	\label{tab:table4}
\end{table}

As it can be seen in the table \ref{tab:table4}, the evaluation of the answer sheet for all the sample cases was very accurate. Although it isn’t that close to what manual evaluation consisted of but if we convert the system into percentile based statistical marking system, the score then will be very close to each other. In case, if the institute wants to rank the students depending on the score, the ranking from the manual and automated system, will appear to be same. Table \ref{tab:table5} shows a general comparison of the proposed model with the models used in earlier researches as discussed in the literature review. Authors have used mean-square-error as one of the metrics to carry out the comparison. The error (E) can be defined as follows:

\centerline{E= 1/m (Summation(i=1 to m) (AE-ME)\^2)}

Where, E is the normalized error \newline
‘m’ is the number of students or cases \newline
‘AE’ is the score generated by Automatic System \newline
‘ME’ is the score assigned by Manual Evaluation 

\begin{table}
	\caption{General comparison of the model  performance with already existing models in the literature.}
	\centering
	\begin{tabular}{lll}
		\toprule
		\multicolumn{3}{c}{}                   \\
		\cmidrule(r){1-3}
		Model     & Algorithm used & Error \\
		\midrule
		Model 1 & RNN & 2.696 \\
		Model 2 & Cosine vector similarity & 1.667 \\
		Model 3 & Jaccard's similarity & 1.312 \\
		Proposed Model & Siamese Manhattan LSTM algorithm & 1.372\\
		\bottomrule
	\end{tabular}
	\label{tab:table5}
\end{table}

\section{Conclusion}
Thus, with advancements in Machine Learning and Computational Intelligence, there are forthcoming system helping to improve the current educational system. This paper presented one such smart automated system for evaluation of descriptive answer based on certain ideal factors fed by the examiner. The system performed very well when implemented as per the results were shown in the paper. The concept used was based on text analysis and text summarization which comes under the field of Natural Language Processing. 

The current model fails to evaluate the answer if the answer consists of figures, diagram, equations, numerical or any such mathematical or pictorial representations. The future objective of this research would be to work upon these limitations of the model. Moreover, authors are working to deploy an advanced Optical Character Recognition (OCR) \citep{sharma2013data} tool in the flow to even evaluate handwritten answer sheets by scanning them into the system.  Authors are also working to extend this research, where the model shall also give analysis of what mistakes the student did on the student dashboard at time of result declaration

\bibliographystyle{unsrtnat}
\bibliography{references}

\end{document}